# A Method for Tumor Treating Fields Fast Estimation

Reuben R Shamir[1][0000-0001-6263-6889] and Zeev Bomzon[1][0000-0002-7813-2252]

[1] Novocure, Israel
rshamir@novocure.com

**Abstract.** Tumor Treating Fields (TTFields) is an FDA approved treatment for specific types of cancer and significantly extends patients' life. The intensity of the TTFields within the tumor was associated with the treatment outcomes: the larger the intensity the longer the patients are likely to survive. Therefore, it was suggested to optimize TTFields transducer array location such that their intensity is maximized. Such optimization requires multiple computations of TTFields in a simulation framework. However, these computations are typically performed using finite element methods or similar approaches that are time consuming. Therefore, only a limited number of transducer array locations can be examined in practice. To overcome this issue, we have developed a method for fast estimation of TTFields intensity. We have designed and implemented a method that inputs a segmentation of the patient's head, a table of tissues' electrical properties and the location of the transducer array. The method outputs a spatial estimation of the TTFields intensity by incorporating a few relevant parameters in a random-forest regressor. The method was evaluated on 10 patients (20 TA layouts) in a leave-one-out framework. The computation time was 1.5 minutes using the suggested method, and 180-240 minutes using the commercial simulation. The average error was 0.14 V/cm (SD = 0.06 V/cm) in comparison to the result of the commercial simulation. These results suggest that a fast estimation of TTFields based on a few parameters is feasible. The presented method may facilitate treatment optimization and further extend patients' life.

**Keywords:** Tumor Treating Fields, Treatment Planning, Simulation.

## 1 Introduction

Tumor Treating Fields (TTFields) therapy is an FDA approved treatment for Glioblastoma Multiforme (GBM) and Malignant Pleural Mesothelioma (MPM) [1, 2]. Clinical trials have shown that adding TTFields to standard of care significantly extends Gllioblastoma patient overall survival [1]. Similar improvements were observed in MPM patients [2]. TTFields are delivered non-invasively using pairs of transducer arrays that are placed on the skin in close proximity to the tumor. The arrays are connected to a field generator that when activated generates an alternating electric field in the range of 100-200 KHz that propagates into the cancerous tissue.

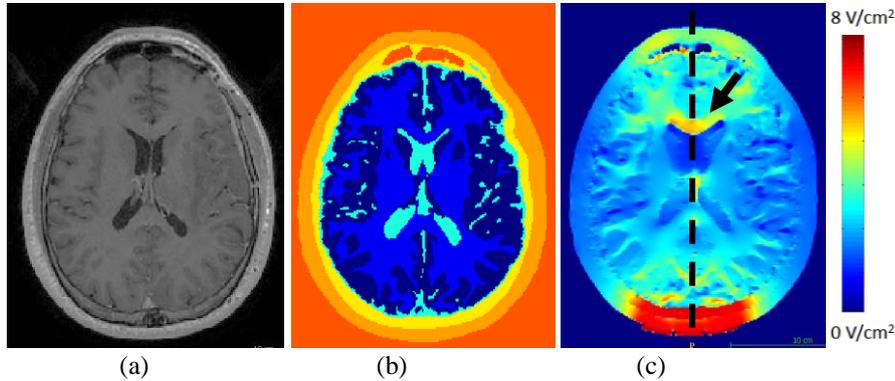

**Fig. 1.** (a) Head MRI T1 with gadolinium of a GBM patient who underwent TTField treatment. (b) Segmentation of the patient's MRI into tissues with different electrical properties. (c) TTfields spatial distribution that was computed with a finite element methods. Note that the TTFields are increased in the vicinity of cerebrospinal fluid (arrow). Moreover, TTFields are larger in tissues that are closer to the dashed line between the centers of TA pairs.

Recent post-hoc analysis of clinical data showed that delivery of higher field intensities to the tumor is associated with prolonged patient survival. [3]. Therefore, placing the transducer arrays such that the TTFields intensity is maximized in the cancerous tissue, has the potential of further extending patients' life.

Finding the array placement that maximizes field intensity in the tumor is an optimization problem that requires calculating the electric field distribution generated by multiple positions of the arrays. Current methods for estimating TTFields intensity distributions rely on finite element methods that are time consuming and may require hours to compute the field generated by a single pair of arrays [4]. Hence, during any practical optimization scheme for TTFields treatment planning, only a limited number of transducer array locations can be evaluated and the optimization result may be suboptimal.

Recent studies have suggested that machine learning methods can be utilized to approximate finite element methods output [4–11]. Benuzzi et al. [8] predicted some mechanical properties in the assembly of a railway axle and wheel. Wang et al. [9] used a regression model to predict road sweeping brush load characteristics. Lostado et al. [11] utilized regression trees to define stress models and later on [10] to determine the maximum load capacity in tapered roller bearings. Guo et al. [12] have proposed to utilize a convolutional neural network for real-time prediction of non-uniform steady laminar flow. Pfeiffer et al. [13] incorporated a convolutional neural network to estimate the spatial displacement of an organ as a response to mechanical forces. Liang et al. [14] have incorporated a deep neural network to estimate biomechanical stress distribution as a fast and accurate surrogate of finite-element methods. Finally, Hennigh et al. [15]

have introduced Lat-Net, a method for compressing both the computation time and memory usage of Lattice Boltzmann flow simulations using deep neural networks.

In this study we present a novel method that incorporates the random forest regression for the fast estimation of TTFields. To the best of our knowledge, it is the first attempt to utilize a machine learning method for significantly decreasing the computation time of TTFields simulation. The key contributions of this study are as follows: 1) identification of key parameters that effect TTFields intensity; 2) a method for extraction of these parameters; 3) utilization of random forest regression for fast estimation of the TTFields, and; 4) validation of the method on 10 GBM patients.

## 2 Methods

### 2.1 Key parameters that effect TTFields

Based on Ohm's law, Maxwell's equations in matter and Coulomb's law, the electric field is inversely related to conductivity ($\sigma$), permittivity ($\varepsilon$), and distance from electrical source ($d_e$), respectively. A close inspection of simulation results (Fig. 1) suggests that the TTFields are larger when the tissue is in the proximity of the cerebrospinal fluid (CSF). A possible explanation for this observation is that electrons are accumulated on the CSF's boundary since of its high conductivity, therefore, increasing the electric potential in these zones. We denote the shortest distance of a voxel from a voxel of CSF as $d_c$. Another observation is that the TTFields are larger in tissues that are closer to the imaginary line between the centers of TA pairs (Fig. 1). This observation is in line with a generalization of Coulomb's law to finite parallel plates in homogenous matter. We denote the distance between a voxel and the line along TA centers as $d_l$. The conductivity and permittivity are expected to have a linear relation with the electric field, and the distance is polynomial to the electric field. Yet, we are unfamiliar with a formula that combines all of the above features, and therefore, incorporate a regression method.

Given patient's head MRI the above key parameters were extracted as follows. At first, we have segmented the head into eight tissues (Fig. 1b): 1) skin and muscle (as one tissue); 2) skull; 3) CSF; 4) white matter; 5) grey matter; 6) tumor – enhancing; 7) tumor – necrotic, and; 8) tumor resection cavity. The segmentation of the tumor was performed semi-automatically using region growing and active contours methods [6]. The segmentation of the head tissues (1-5) was performed automatically with a custom atlas-based method [16]. The conductivity and permittivity of the different tissues were determined as described in [17]. The distances of each voxel from electrical source, CSF and the line along TA centers were efficiently computed using the method presented in Danielsson et al. [5].

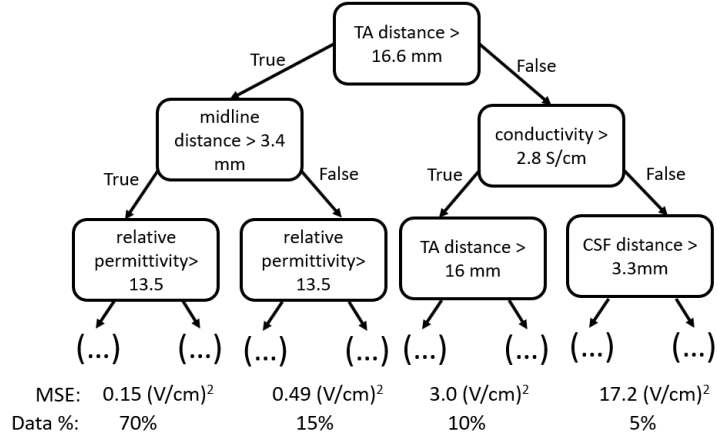

**Fig. 2.** The first three layers of a decision tree regressor that was trained to predict the tumor treating fields strength. The mean squared error (MSE) was reduced for locations that are further from the transducer array (TA): compare the left and right branches' accuracies in this example.

### 2.2 Random forests regression for estimation of TTFields

Random forests are an ensemble of decision tree predictors, such that each tree is restricted by a random vector that governs the sensitivity of the tree to the input features [18]. Lostado et al. [11] have demonstrated that regression trees facilitate effective modeling of FEM-based non-linear maps for fields of mechanical force. Moreover, they suggest that since random forests divide the dataset into groups of similar features and facilitate local group fitting, good models can be generated also when the data is heterogeneous, irregular, and of limited size.

Therefore, we have incorporated a random forest regressor. We set up 30 trees, mean squared error quality of split measure, using bootstrap and out-of-bag samples to estimate regression quality on unseen samples. The number of trees was selected by a trial-and-error process to balance accuracy and prediction-time tradeoff. The input per voxel to the regression tree is as follows. 1) conductivity ($\sigma$); 3) permittivity ($\varepsilon$); 4) distance from closest electrical source ($d_e$); 5) distance from closest CSF ($d_c$), and; 6) distance from TAs midline ($d_l$).

We investigated the relevance of the above features to the prediction in our experimental setup (see below) using mean decrease in impurity method that results with a feature's importance score in the range of 0 to 1 [18]. The distance of closest electrical source was by far the most important feature (0.65). Distance from TA midline and from CSF were of secondary importance (0.15 and 0.1, respectively). Conductivity and permittivity importance scores were both 0.05. A typical example for a decision tree in the random forest is presented in Fig. 2. An interesting observation is that the mean squared error (MSE) was reduced for locations that are further from the TA. The specific tree in this example splits the data at 16.6mm distance to TA. Indeed, larger errors were observed for the 15% of data that are within this range.

4**Table 1.** Experiment results. The average (SD) of absolute differences between the TTFields computed with a finite element method and the random forest. Results for pairs of transducer arrays along the anterior-posterior (AP) and left-right (LR) axes of the head are presented.

| Patient # | AP Error (V/cm) | LR Error (V/cm) | Patient # | AP Error (V/cm) | LR Error (V/cm) |
|---|---|---|---|---|---|
| 1 | 0.14 (0.62) | 0.23 (0.80) | 6 | 0.08 (0.45) | 0.08 (0.43) |
| 2 | 0.16 (0.56) | 0.19 (0.71) | 7 | 0.15 (0.61) | 0.14 (0.62) |
| 3 | 0.15 (0.69) | 0.17 (0.59) | 8 | 0.10 (0.50) | 0.11 (0.49) |
| 4 | 0.15 (0.61) | 0.18 (0.66) | 9 | 0.14 (0.60) | 0.14 (0.64) |
| 5 | 0.15 (0.61) | 0.15 (0.59) | 10 | 0.11 (0.55) | 0.13 (0.62) |

In addition, we compared the random forest to a multi-linear regression. Specifically, the following linear formula was incorporated to estimate the TTFields.

$$|E| \sim a_0 + a_1 \sigma^{-1} + a_2 \varepsilon^{-1} + a_3 d_e^{-1} + a_4 d_e^{-2} + a_5 d_c + a_6 d_l \quad (1)$$

The coefficients $a_i$ were computed to best fit the finite elements method output to the linear regression model (see next section).

### 2.3 Experimental setup

We have validated the suggested method using a dataset of 10 patients that underwent TTFields therapy. At first, the patients' MRIs were segmented as described in Section 2.1 and the head's outer surface was extracted using the marching cubes algorithm [19]. Then, two TA pairs were virtually placed on the head's surface. The first pair was placed such that one TA is on the forehead and the other one is on the back of the head. In this case, the TTField direction is roughly parallel to the anterior-posterior (AP) axis. The second pair was placed such that the TAs are on opposite lateral sides of the head. In this case, the TTField direction is roughly parallel to the left-right (LR) axis of the head. For each of the 20 pairs, we computed the absolute electric field intensity spatial distribution with a finite element method (Fig. 3). That is, we associate the electric field for each voxel in the patient's MRI. We marked this dataset as gold standard as it was verified in phantoms and associated with patients' survival [3].

We use a leave-one-out approach for the training. One test patient was excluded at a time while the 18 datasets of the rest nine patients were incorporated to train the random forest and the multilinear regression model (Eq. 1). Then, the TTFields were predicted using random forest and multilinear regression model on the test patient data. Large parts of the image are associated with air that is not conductive. Therefore, it can bias the result of the model. To handle this situation, we consider only a small portion of the voxels with air in the training by ensuring that their number is similar to those in other segmented tissues. In this study, the training and prediction are performed per voxel independently of its neighbors.

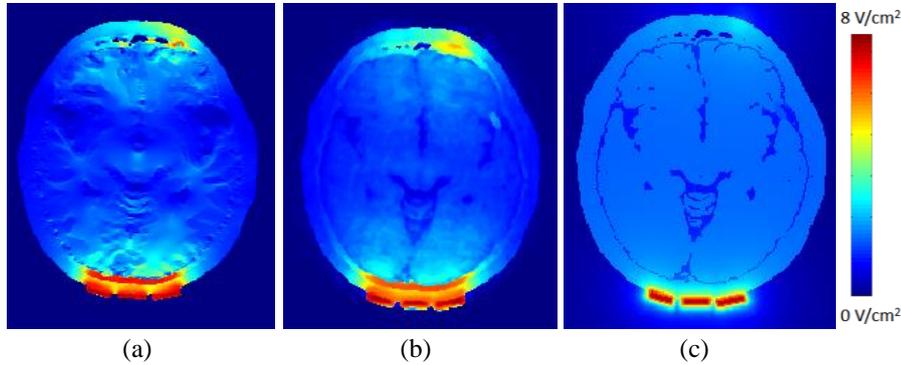

|     |     |     |
| :-: | :-: | :-: |
| (a) | (b) | (c) |

**Fig. 3.** TTFields estimation that was computed by the gold standard finite-elements method (a), random forest (b), and linear regression (c).

We have implemented the suggested method with Python 3.6 using scipy [20], numpy [21], scikit-learn [22] and SimpleITK [23] packages. We used 3D Slicer [24] for visual inspection of the results. The method was executed on a standard desktop computer (Intel i7 CPU, 16 GB RAM) with Windows 10 operating system (Microsoft, Redmond, WA, USA). The gold standard simulations were computed using sim4life (Zurich Med Tech, Zurich, Switzerland) on a dedicated simulation computer (Intel i7 CPU, NVidia 1080 Ti GPU, 128 GB RAM) with Windows 10 operating system (Microsoft, Redmond, WA, USA). Patients' MRIs were T1 weighted with gadolinium with voxel spacing of 1x1x1 mm$^3$ and incorporated the entire head.

## 3     Results

The average absolute differences between the random forest prediction and the gold standard was 0.14 V/cm (patients' SD = 0.035, range 0.08 – 0.23 V/cm, N = 20). Table 1 presents a per-patient summary of our results. The random forest resulted with a better accuracy in comparison to the multilinear regression. Compare the random forest average absolute differences above and the linear regression results of 0.29 V/cm (patients' SD = 0.04, range 0.23 – 0.37 V/cm, N = 20). The random forest average prediction time was 15 seconds (SD = 1.5 seconds). Note that this measure is excluding the preprocessing required to extract the distance measures that typically required 30 seconds for each: TA, CSF and midline.

Fig. 3 presents typical electric-field spatial distributions that were computed by the gold-standard, the random forest, and the multilinear regression. Fig. 4 demonstrates typical absolute differences between the gold-standard and random-forest prediction. The values are very similar (<0.4 V/cm) in most locations. However, large errors (> 2 V/cm) were observed in the vicinity of the TAs (Fig. 4a) and instantly outside the ventricles along the TA main axis (Fig. 4b).

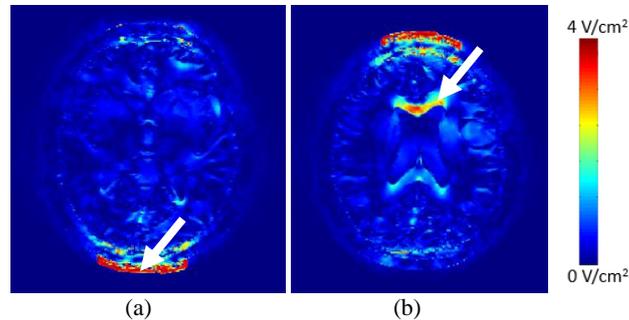

(a)          (b)

**Fig. 4.** Absolute differences between gold standard finite-elements method and random-forest based estimation. Larger errors were observed near the transducer arrays (a) and near the ventricles along the main transducer array axis (b)

## 4   Discussion

We have presented a novel method for the fast estimation of TTFields spatial distribution and demonstrated that average accuracy of 0.14 V/cm can be achieved within a short time. Compare the ~1.5 minutes computation time with the suggested random forest method to the 3-4 hours computation time using the gold standard method. Note that the computation time can be further reduced by a factor of three by the parallelization of data preparation.

Selection of optimal TA location involves the computation of average TTFields over a tumor area. Averaging is expected to further improve accuracy. Yet, the utilization of the random forest method TTFields estimation for optimization of TA placement is out of the scope of this study and requires further investigation.

One limitation of our method is that it assumes that the effect of neighbor voxels is minor and can be neglected. We plan to revise our method to incorporate also neighbor voxels and consider convolutional and recurrent neural networks. Another limitation is the data-preparation computation time. We plan to investigate one-time distance-maps computation and fast manipulation of these maps upon alternation of TA locations to reduce overall computation time to a few seconds. Last, we plan to extend the method to the chest and abdomen for supporting additional indications of TTFields treatment.